\documentclass[11pt,a4paper]{article}
\usepackage{authblk}
\usepackage[hyperref]{emnlp2017} 
\usepackage{times}
\usepackage{latexsym}

\usepackage{url}
\usepackage{booktabs}
\usepackage{amsmath}
\usepackage{amssymb}

\usepackage{siunitx}
\usepackage[colorinlistoftodos,prependcaption,textsize=tiny]{todonotes}

\emnlpfinalcopy 

\newcommand{\ndsets}{8}
\newcommand{\ndomains}{5}

\usepackage{array}
\newcolumntype{C}[1]{>{\centering\let\newline\\\arraybackslash\hspace{0pt}}m{#1}}
\newcolumntype{L}[1]{>{\let\newline\\\arraybackslash\hspace{0pt}}m{#1}}

\title{Using millions of emoji occurrences to learn any-domain representations for detecting sentiment, emotion and sarcasm}

\author[1]{Bjarke Felbo}
\author[2]{Alan Mislove}
\author[3]{Anders S{\o}gaard}
\author[1]{Iyad Rahwan}
\author[4]{Sune Lehmann}
\affil[1]{Media Lab, Massachusetts Institute of Technology}
\affil[2]{College of Computer and Information Science, Northeastern University}
\affil[3]{Department of Computer Science, University of Copenhagen}
\affil[4]{DTU Compute, Technical University of Denmark}

\date{}

\begin{document}
\maketitle
\begin{abstract}

NLP tasks are often limited by scarcity of manually annotated data. In social media sentiment analysis and related tasks, researchers have therefore used binarized emoticons and specific hashtags as forms of distant supervision. Our paper shows that by extending the distant supervision to a more diverse set of noisy labels, the models can learn richer representations. Through emoji prediction on a dataset of 1246 million tweets containing one of 64 common emojis we obtain state-of-the-art performance on \ndsets{}~benchmark datasets within sentiment, emotion and sarcasm detection using a single pretrained model. Our analyses confirm that the diversity of our emotional labels yield a performance improvement over previous distant supervision approaches.

\end{abstract}

\section{Introduction}
\label{sec:introduction}

A variety of NLP tasks are limited by scarcity of manually annotated data. Therefore, co-occurring emotional expressions have been used for distant supervision in social media sentiment analysis and related tasks to make the models learn useful text representations before modeling these tasks directly. For instance, the state-of-the-art approaches within sentiment analysis of social media data use positive/negative emoticons for training their models~\cite{deriu2016swisscheese, tang2014learning}. Similarly, hashtags such as \#anger, \#joy, \#happytweet, \#ugh, \#yuck and \#fml have in previous research been mapped into emotional categories for emotion analysis~\cite{mohammad2012emotional}.

Distant supervision on noisy labels often enables a model to obtain better performance on the target task. In this paper, we show that extending the distant supervision to a more diverse set of noisy labels enables the models to learn richer representations of emotional content in text, thereby obtaining better performance on benchmarks for detecting sentiment, emotions and sarcasm. We show that the learned representation of a single pretrained model generalizes across \ndomains{} domains.

\begin{table}[htp]
\caption{Example sentences scored by our model. For each text the top five most likely emojis are shown with the model's probability estimates.}
\label{tab:sentence_list}
\begin{center}
\includegraphics[trim=0.05cm 0.0cm 0.0cm 0.0cm, clip, width=1\columnwidth]{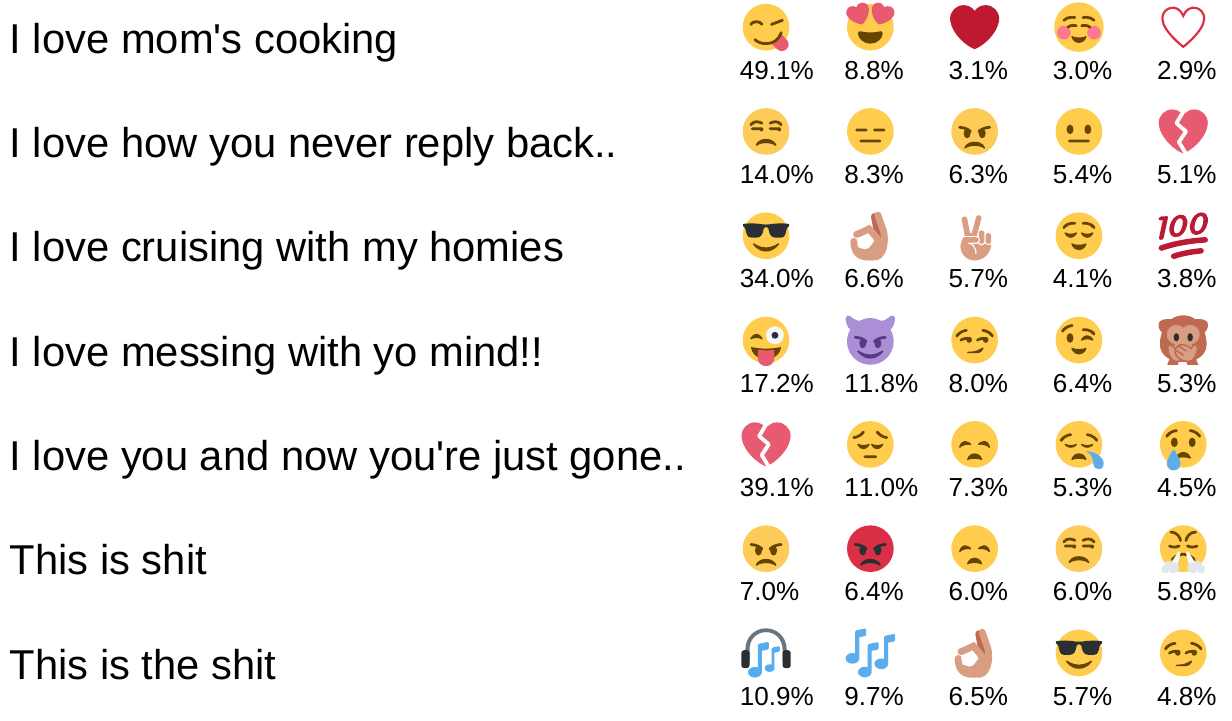}
\end{center}
\end{table}

Emojis are not always a direct labeling of emotional content. For instance, a positive emoji may serve to disambiguate an ambiguous sentence or to complement an otherwise relatively negative text. ~\citet{kunneman2014predictability} discuss a similar duality in the use of emotional hashtags such as {\em \#nice}~and {\em \#lame}. Nevertheless, our work shows that emojis can be used to classify the emotional content of texts accurately in many cases. For instance, our DeepMoji model captures varied usages of the word `love' as well as slang such as `this is the shit' being a positive statement (see Table~\ref{tab:sentence_list}). We provide an online demo at deepmoji.mit.edu to allow others to explore the predictions of our model. 

\paragraph{Contributions} 
We show how millions of readily available emoji occurrences on Twitter can be used to pretrain models to learn a richer emotional representation than traditionally obtained through distant supervision. We transfer this knowledge to the target tasks using a new layer-wise fine-tuning method, obtaining improvements over the state-of-the-art within a range of tasks: emotion, sarcasm and sentiment detection. We present multiple analyses on the effect of pretraining, including results that show that the diversity of our emoji set is important for the transfer learning potential of our model. Our pretrained DeepMoji model is released with the hope that other researchers can use it for various NLP tasks\footnote{Available with preprocessing code, examples of usage, benchmark datasets etc. at github.com/bfelbo/deepmoji}.

\section{Related work}
\label{sec:related_work}

Using emotional expressions as noisy labels in text to counter scarcity of labels is not a new idea~\cite{read2005using, go2009twitter}. Originally, binarized emoticons were used as noisy labels, but later also hashtags and emojis have been used. To our knowledge, previous research has always manually specified which emotional category each emotional expression belong to. Prior work has used theories of emotion such as Ekman's six basic emotions and Plutchik's eight basic emotions~\cite{mohammad2012emotional,suttles2013}.

Such manual categorization requires an understanding of the emotional content of each expression, which is difficult and time-consuming for sophisticated combinations of emotional content. Moreover, any manual selection and categorization is prone to misinterpretations and may omit important details regarding usage. In contrast, our approach requires no prior knowledge of the corpus and can capture diverse usage of 64 types of emojis (see Table~\ref{tab:sentence_list} for examples and Figure~\ref{fig:tree} for how the model implicitly groups emojis).

Another way of automatically interpreting the emotional content of an emoji is to learn emoji embeddings from the words describing the emoji-semantics in official emoji tables~\cite{eisner2016emoji2vec}. This approach, in our context, suffers from two severe limitations: a) It requires emojis at test time while there are many domains with limited or no usage of emojis. b) The tables do not capture the dynamics of emoji usage, i.e., drift in an emoji's intended meaning over time.

Knowledge can be transferred from the emoji dataset to the target task in many different ways. In particular, multitask learning with simultaneous training on multiple datasets has shown promising results~\cite{collobert2008unified}. However, multitask learning requires access to the emoji dataset whenever the classifier needs to be tuned for a new target task. Requiring access to the dataset is problematic in terms of violating data access regulations. There are also issues from a data storage perspective as the dataset used for this research contains hundreds of millions of tweets (see Table~\ref{tab:emoji_matrix}). Instead we use transfer learning~\cite{bengio2012deep} as described in \S\ref{sub_sec:transfer_learning}, which does not require access to the original dataset, but only the pretrained classifier.

\section{Method}
\label{sec:method}

\subsection{Pretraining}
\label{sub_sec:pretraining}

\begin{figure}[tp]
  \centering
  \includegraphics[trim=0.0cm 0.0cm 0.0cm 0.0cm, clip, width=0.5\columnwidth]{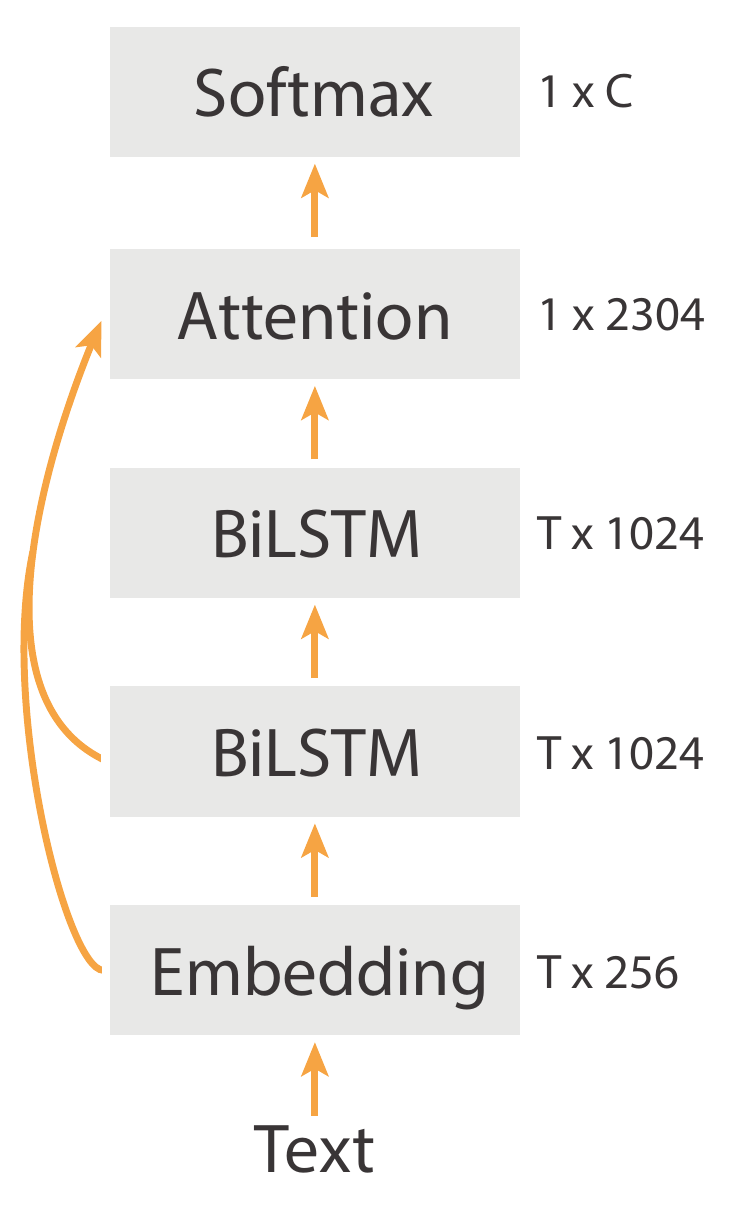}
  \caption{Illustration of the DeepMoji model with $T$ being text length and $C$ the number of classes.}
  \label{fig:lstm}
\end{figure}

In many cases, emojis serve as a proxy for the emotional contents of a text. Therefore, pretraining on the classification task of predicting which emoji were initially part of a text can improve performance on the target task (see \S\ref{sub_sec:understanding_pretraining} for an analysis of why our pretraining helps). Social media contains large amounts of short texts with emojis that can be utilized as noisy labels for pretraining. Here, we use data from Twitter from January 1st 2013 to June 1st 2017, but any dataset with emoji occurrences could be used.

Only English tweets without URL's are used for the pretraining dataset. Our hypothesis is that the content obtained from the URL is likely to be important for understanding the emotional content of the text in the tweet. Therefore, we expect emojis associated with these tweets to be noiser labels than for tweets without URLs, and the tweets with URLs are thus removed.

Proper tokenization is important for generalization. All tweets are tokenized on a word-by-word basis. Words with 2 or more repeated characters are shortened to the same token (e.g. `loool' and `looooool' are tokenized such that they are treated the same). Similarly, we use a special token for all URLs (only relevant for benchmark datasets), user mentions (e.g. `@acl2017' and `@emnlp2017' are thus treated the same) and numbers. To be included in the training set the tweet must contain at least 1 token that is not a punctuation symbol, emoji or special token\footnote{Details available at github.com/bfelbo/deepmoji}.

Many tweets contain multiple repetitions of the same emoji or multiple different emojis. In the training data, we address this in the following way. For each unique emoji type, we save a separate tweet for the pretraining with that emoji type as the label. We only save a single tweet for the pretraining per unique emoji type regardless of the number of emojis associated with the tweet. This data preprocessing allows the pretraining task to capture that multiple types of emotional content are associated with the tweet while making our pretraining task a single-label classification instead of a more complicated multi-label classification.

To ensure that the pretraining encourages the models to learn a rich understanding of emotional content in text rather than only emotional content associated with the most used emojis, we create a balanced pretraining dataset. The pretraining data is split into a training, validation and test set, where the validation and test set is randomly sampled in such a way that each emoji is equally represented. The remaining data is upsampled to create a balanced training dataset.

\subsection{Model}
\label{sub_sec:model}

With the millions of emoji occurrences available, we can train very expressive classifiers with limited risk of overfitting. We use a variant of the Long Short-Term Memory (LSTM) model that has been successful at many NLP tasks~\cite{hochreiter1997long, sutskever2014sequence}. Our DeepMoji model uses an embedding layer of 256 dimensions to project each word into a vector space. A hyperbolic tangent activation function is used to enforce a constraint of each embedding dimension being within $[-1, 1]$. To capture the context of each word we use two bidirectional LSTM layers with 1024 hidden units in each (512 in each direction). Finally, an attention layer that take all of these layers as input using skip-connections is used (see Figure~\ref{fig:lstm} for an illustration).

The attention mechanism lets the model decide the importance of each word for the prediction task by weighing them when constructing the representation of the text. For instance, a word such as `amazing' is likely to be very informative of the emotional meaning of a text and it should thus be treated accordingly. We use a simple approach inspired by~\cite{bahdanau2014neural, yang2016hierarchical} with a single parameter pr. input channel:

\begin{align*}
e_t &= h_t w_a\\
a_t &= \frac{exp(e_t)}{\sum_{i=1}^{T}{exp(e_i)}}\\
v &= \sum_{i=1}^{T}{a_i h_i}
\end{align*}

Here $h_t$ is the representation of the word at time step $t$ and $w_a$ is the weight matrix for the attention layer. The attention importance scores for each time step, $a_t$, are obtained by multiplying the representations with the weight matrix and then normalizing to construct a probability distribution over the words. Lastly, the representation vector for the text, $v$, is found by a weighted summation over all the time steps using the attention importance scores as weights. This representation vector obtained from the attention layer is a high-level encoding of the entire text, which is used as input to the final Softmax layer for classification. We find that adding the attention mechanism and skip-connections improves the model's capabilities for transfer learning (see \S\ref{sub_sec:analysis_model_architecture} for more details). 

The only regularization used for the pretraining task is a L2 regularization of \num{1E-6} on the embedding weights. For the finetuning additional regularization is applied (see \S\ref{sub_sec:benchmarking}). Our model is implemented using Theano~\cite{theano} and we make an easy-to-use version available that uses Keras~\cite{chollet2015keras}.

\subsection{Transfer learning}
\label{sub_sec:transfer_learning}

Our pretrained model can be fine-tuned to the target task in multiple ways with some approaches `freezing' layers by disabling parameters updates to prevent overfitting. One common approach is to use the network as a feature extractor~\cite{donahue2014decaf}, where all layers in the model are frozen when fine-tuning on the target task except the last layer (hereafter referred to as the \textit{`last'} approach). Alternatively, another common approach is to use the pretrained model as an initialization~\cite{erhan2010does}, where the full model is unfrozen (hereafter referred to as \textit{`full'}).

We propose a new simple transfer learning approach, \textit{`chain-thaw'}, that sequentially unfreezes and fine-tunes a single layer at a time. This approach increases accuracy on the target task at the expense of extra computational power needed for the fine-tuning. By training each layer separately the model is able to adjust the individual patterns across the network with a reduced risk of overfitting. The sequential fine-tuning seems to have a regularizing effect similar to what has been examined with layer-wise training in the context of unsupervised learning~\cite{erhan2010does}.

\begin{figure}[tp]
  \centering
  \includegraphics[trim=0.3cm 0.0cm 0.3cm 0.0cm, clip, width=1\columnwidth]{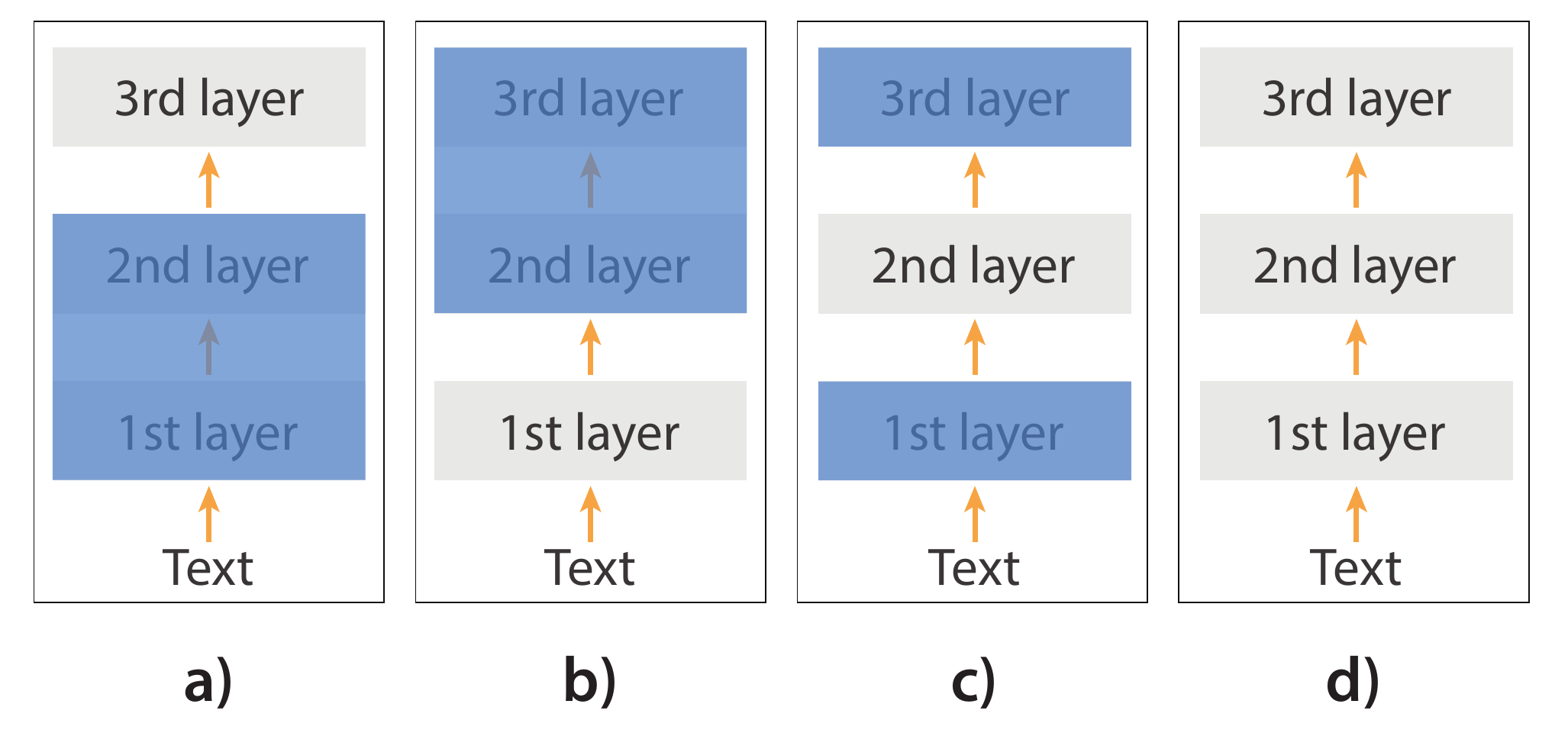}
  \caption{Illustration of the chain-thaw transfer learning approach, where each layer is fine-tuned separately. Layers covered with a blue rectangle are frozen. Step a) tunes any new layers, b) then tunes the 1st layer and c) the next layer until all layers have been fine-tuned individually. Lastly, in step d) all layers are fine-tuned together.}
  \label{fig:chainthaw}
\end{figure}

More specifically, the chain-thaw approach first fine-tunes any new layers (often only a Softmax layer) to the target task until convergence on a validation set. Then the approach fine-tunes each layer individually starting from the first layer in the network. Lastly, the entire model is trained with all layers. Each time the model converges as measured on the validation set, the weights are reloaded to the best setting, thereby preventing overfitting in a similar manner to early stopping~\cite{sjoberg1995overtraining}. This process is illustrated in Figure~\ref{fig:chainthaw}. Note how only performing step a) in the figure is identical to the `last' approach, where the existing network is used as a feature extractor. Similarly, only doing step d) is identical to the `full' approach, where the pretrained weights are used as an initialization for a fully trainable network. Although the chain-thaw procedure may seem extensive it is easily implemented with only a few lines of code. Similarly, the additional time spent on fine-tuning is limited when the target task uses GPUs on small datasets of manually annotated data as is often the case.

A benefit of the chain-thaw approach is the ability to expand the vocabulary to new domains with little risk of overfitting. For a given dataset up to 10000 new words from the training set are added to the vocabulary. \S\ref{sub_sec:understanding_pretraining} contains analysis on the added word coverage gained from this approach. 

\begin{table}[htp]
\caption{The number of tweets in the pretraining dataset associated with each emoji in millions.}
\label{tab:emoji_matrix}
\begin{center}
\includegraphics[trim=0.0cm 0.0cm 0.0cm 0.0cm, clip, width=1\columnwidth]{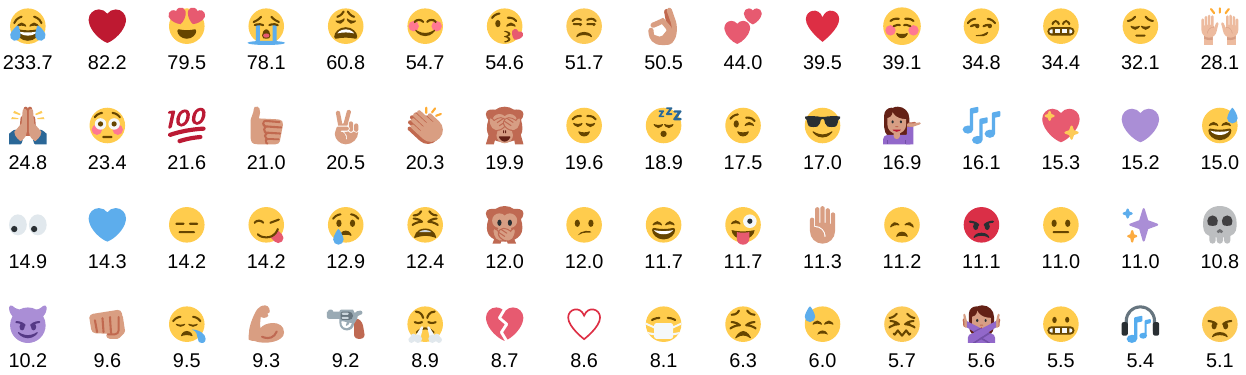}
\end{center}
\end{table}

\begin{table}[h]
\centering
\small
\caption{Accuracy of classifiers on the emoji prediction task. $d$ refers to the dimensionality of each LSTM layer. Parameters are in millions.}
\label{tab:pretrain_results}
\begin{center}
\begin{tabular}{@{\hspace{3pt}}l@{\hspace{3pt}}ccc}
\toprule
\hspace{2.2cm} & Params &  Top 1 & Top 5  \\
 \midrule
 Random  & $-$ & $1.6\%$ & $7.8\%$  \\
 fasttext & $12.8$ & $12.8\%$ & $36.2\%$ \\ 
 DeepMoji (d = 512) & $15.5$ & $16.7\%$ & $43.3\%$ \\
 DeepMoji (d = 1024) & $22.4$ & $17.0\%$ & $43.8\%$ \\
\bottomrule
\end{tabular}
\end{center}
\end{table}

\section{Experiments}
\label{sec:experiments}

\subsection{Emoji prediction}
\label{sub_sec:emoji_prediction}

We use a raw dataset of $56.6$ billion tweets, which is then filtered to $1.2$ billion relevant tweets (see details in \S\ref{sub_sec:pretraining}). In the pretraining dataset a copy of a single tweet is stored once for each unique emoji, resulting in a dataset consisting of 1.6 billion tweets. Table~\ref{tab:emoji_matrix} shows the distribution of tweets across different emoji types. To evaluate performance on the pretraining task a validation set and a test set both containing $640$K tweets ($10$K of each emoji type) are used. The remaining tweets are used for the training set, which is balanced using upsampling.

\begin{table*}[h]
\centering
\small
\caption{Description of benchmark datasets. Datasets without pre-existing training/test splits are split by us (with splits publicly available). Data used for hyperparameter tuning is taken from the training set.}
\label{tab:benchmark_dsets}
\begin{center}
\begin{tabular}{lcccccc}
\toprule
Identifier & Study & Task & Domain & Classes  & $N_{train}$ & $N_{test}$ \\
 \midrule
 SE0714  & \cite{strapparava2007semeval} & Emotion & Headlines & $3$ & $250$ & $1000$ \\
   Olympic  & \cite{sintsova2013fine} & Emotion & Tweets & $4$  & $250$ & $709$ \\
  PsychExp  & \cite{wallbott1986universal} & Emotion & Experiences & $7$ & $1000$ & $6480$ \\
 \midrule
 SS-Twitter  & \cite{thelwall2012sentiment} & Sentiment & Tweets & $2$ &  $1000$ & $1113$ \\
 SS-Youtube  & \cite{thelwall2012sentiment} & Sentiment & Video Comments & $2$ &  $1000$ & $1142$ \\
 SE1604  & \cite{nakov2016semeval} & Sentiment & Tweets & $3$ & $7155$ & $31986$ \\
 \midrule
 SCv1 & \cite{walker2012corpus} & Sarcasm & Debate Forums & $2$ & $1000$ & $995$ \\ 
  SCv2-GEN & \cite{oraby2016creating} & Sarcasm & Debate Forums & $2$ & $1000$ & $2260$ \\ 
\bottomrule
\end{tabular}
\end{center}
\end{table*}

The performance of the DeepMoji model is evaluated on the pretraining task with the results shown in Table~\ref{tab:pretrain_results}. Both top 1 and top 5 accuracy is used for the evaluation as the emoji labels are noisy with multiple emojis being potentially correct for any given sentence. For comparison we also train a version of our DeepMoji model with smaller LSTM layers and a bag-of-words classifier, fastText, that has recently shown competitive results~\cite{joulin2016bag}. We use 256 dimensions for this fastText classifier, thereby making it almost identical to only using the embedding layer from the DeepMoji model. The difference in top 5 accuracy between the fastText classifier ($36.2\%$) and the largest DeepMoji model ($43.8\%$) underlines the difficulty of the emoji prediction task. As the two classifiers only differ in that the DeepMoji model has LSTM layers and an attention layer between the embedding and Softmax layer, this difference in accuracy demonstrates the importance of capturing the context of each word. 

\subsection{Benchmarking}
\label{sub_sec:benchmarking}

We benchmark our method on 3 different NLP tasks using \ndsets{} datasets across \ndomains{} domains. To make for a fair comparison, we compare DeepMoji to other methods that also utilize external data sources in addition to the benchmark dataset. An averaged F1-measure across classes is used for evaluation in emotion analysis and sarcasm detection as these consist of unbalanced datasets while sentiment datasets are evaluated using accuracy.

An issue with many of the benchmark datasets is data scarcity, which is particularly problematic within emotion analysis. Many recent papers proposing new methods for emotion analysis such as \cite{staiano2014depechemood} only evaluate performance on a single benchmark dataset, SemEval 2007 Task 14, that contains 1250 observations. Recently, criticism has been raised concerning the use of correlation with continuous ratings as a measure~\cite{sven_emotions}, making only the somewhat limited binary evaluation possible. We only evaluate the emotions \{Fear, Joy, Sadness\} as the remaining emotions occur in less than 5\% of the observations.

To fully evaluate our method on emotion analysis against the current methods we thus make use of two other datasets: A dataset of emotions in tweets related to the Olympic Games created by \citeauthor{sintsova2013fine} that we convert to a single-label classification task and a dataset of self-reported emotional experiences created by a large group of psychologists~\cite{wallbott1986universal}. See the supplementary material for details on the datasets and the preprocessing. As these two datasets do not have prior evaluations, we evaluate against a state-of-the-art approach, which is based on a valence-arousal-dominance framework~\cite{sven_emotions}. The scores extracted using this approach are mapped to the classes in the datasets using a logistic regression with parameter optimization using cross-validation. We release our preprocessing code and hope that these 2 two datasets will be used for future benchmarking within emotion analysis.

\begin{table*}[h]
\centering
\small
\caption{Comparison across benchmark datasets. Reported values are averages across five runs. Variations refer to transfer learning approaches in \S\ref{sub_sec:transfer_learning} with `new' being a model trained without pretraining.}
\label{tab:benchmark_results}
\begin{center}
\begin{tabular}{lC{1.5cm}L{2.1cm}C{1.8cm}C{1.8cm}C{1.8cm}C{1.8cm}}
\toprule
Dataset & Measure & State of the art & DeepMoji (new) & DeepMoji (full) & DeepMoji (last) & DeepMoji (chain-thaw)\\
\midrule
 SE0714  &  F1 & $.34$~[Buechel] & $.21$ & $.31$ & $.36$ & $\mathbf{.37}$ \\
 Olympic  &  F1 & $.50$~[Buechel] & $.43$ & $.50$ & $\mathbf{.61}$ & $\mathbf{.61}$ \\
 PsychExp  & F1 & $.45$~[Buechel] & $.32$ & $.42$ & $.56$ & $\mathbf{.57}$ \\
\midrule
 SS-Twitter  & Acc & $.82$~[Deriu] & $.62$ & $.85$ & $.87$ & $\mathbf{.88}$ \\  
 SS-Youtube  & Acc & $.86$~[Deriu] & $.75$ & $.88$ & $.92$ & $\mathbf{.93}$ \\ 
 SE1604  & Acc & $.51$~[Deriu]\footnote{SwissCheese footnote} & $.51$ & $.54$ & $\mathbf{.58}$ & $\mathbf{.58}$ \\ 
 \midrule
 SCv1  & F1 & $.63$~[Joshi] & $.67$ & $.65$ & $.68$ & $\mathbf{.69}$ \\  
 SCv2-GEN  & F1 & $.72$~[Joshi] & $.71$ & $.71$ & $.74$ & $\mathbf{.75}$ \\ 
\bottomrule
\end{tabular}
\end{center}
\end{table*}

\nocite{sven_emotions}
\nocite{deriu2016swisscheese}
\nocite{joshi2016word}

We evaluate sentiment analysis performance on three benchmark datasets. These small datasets are chosen to emphasize the importance of the transfer learning ability of the evaluated models. Two of the datasets are from SentiStrength~\cite{thelwall2010sentiment}, SS-Twitter and SS-Youtube, and follow the relabeling described in \cite{saif2013evaluation} to make the labels binary. The third dataset is from SemEval 2016 Task4A~\cite{nakov2016semeval}. Due to tweets being deleted from Twitter, the SemEval dataset suffers from data decay, making it difficult to compare results across papers. At the time of writing, roughly 15\% of the training dataset for SemEval 2016 Task 4A was impossible to obtain. We choose not to use review datasets for sentiment benchmarking as these datasets contain so many words pr. observation that even bag-of-words classifiers and unsupervised approaches can obtain a high accuracy~\cite{joulin2016bag, radford2017learning}.

The current state of the art for sentiment analysis on social media (and winner of SemEval 2016 Task 4A) uses an ensemble of convolutional neural networks that are pretrained on a private dataset of tweets with emoticons, making it difficult to replicate~\citep{deriu2016swisscheese}. Instead we pretrain a model with the hyperparameters of the largest model in their ensemble on the positive/negative emoticon dataset from~\citet{go2009twitter}. Using this pretraining as an initialization we finetune the model on the target tasks using early stopping on a validation set to determine the amount of training. We also implemented the Sentiment-Specific Word Embedding (SSWE) using the embeddings available on the authors' website~\cite{tang2014learning}, but found that it performed worse than the pretrained convolutional neural network. These results are therefore excluded.

\footnotetext{The authors report a higher accuracy in their paper, which is likely due to having a larger training dataset as they were able to obtain it before data decay occurred.} 

For sarcasm detection we use the sarcasm dataset version 1 and 2 from the Internet Argument Corpus~\cite{walker2012corpus}. Note that results presented on these benchmarks in e.g. \citet{oraby2016creating} are not directly comparable as only a subset of the data is available online.\footnote{We contacted the authors, but were unable to obtain the full dataset for neither version 1 or version 2.} A state-of-the-art baseline is found by modeling the embedding-based features from \citet{joshi2016word} alongside unigrams, bigrams and trigrams with an SVM. GoogleNews word2vec embeddings~\cite{mikolov2013distributed} are used for computing the embedding-based features. A hyperparameter search for regularization parameters is carried out using cross-validation. Note that the sarcasm dataset version 2 contains both a quoted text and a sarcastic response, but to keep the models identical across the datasets only the response is used.

For training we use the Adam optimizer~\cite{kingma2014adam} with gradient clipping of the norm to $1$. Learning rate is set to \num{1E-3} for training of all new layers and \num{1E-4} for finetuning any pretrained layers. To prevent overfitting on the small datasets, $10$\% of the channels across all words in the embedding layer are dropped out during training. Unlike e.g. ~\cite{gal2016theoretically} we do not drop out entire words in the input as some of our datasets contain observations with so few words that it could change the meaning of the text. In addition to the embedding dropout, L2 regularization for the embedding weights is used and $50$\% dropout is applied to the penultimate layer.

Table~\ref{tab:benchmark_results} shows that the DeepMoji model outperforms the state of the art across all benchmark datasets and that our new `chain-thaw' approach consistently yields the highest performance for the transfer learning, albeit often only slightly better or equal to the `last' approach. Results are averaged across 5 runs to reduce the variance. We test the statistical significance of our results by comparing the performance of DeepMoji (chain-thaw) vs. the state of the art. Bootstrap testing with 10000 samples is used. Our results are statistically significantly better than the state of the art with $p < 0.001$ on every benchmark dataset.

Our model is able to out-perform the state-of-the-art on datasets that originate from domains that differ substantially from the tweets on which it was pretrained. A key difference between the pretraining dataset and the benchmark datasets is the length of the observations. The average number of tokens pr. tweet in the pretraining dataset is 11, whereas e.g. the board posts from the Internet Argument Corpus version~1~\cite{oraby2016creating} has an average of 66 tokens with some observations being much longer. 

\section{Model Analysis}
\label{sec:analysis}

\subsection{Importance of emoji diversity}
\label{sub_sec:emoji_diversity}

\begin{figure*}[tp]
  \centering
  \includegraphics[trim=0.0cm 0.0cm 0.0cm 0.0cm, clip, width=1\textwidth]{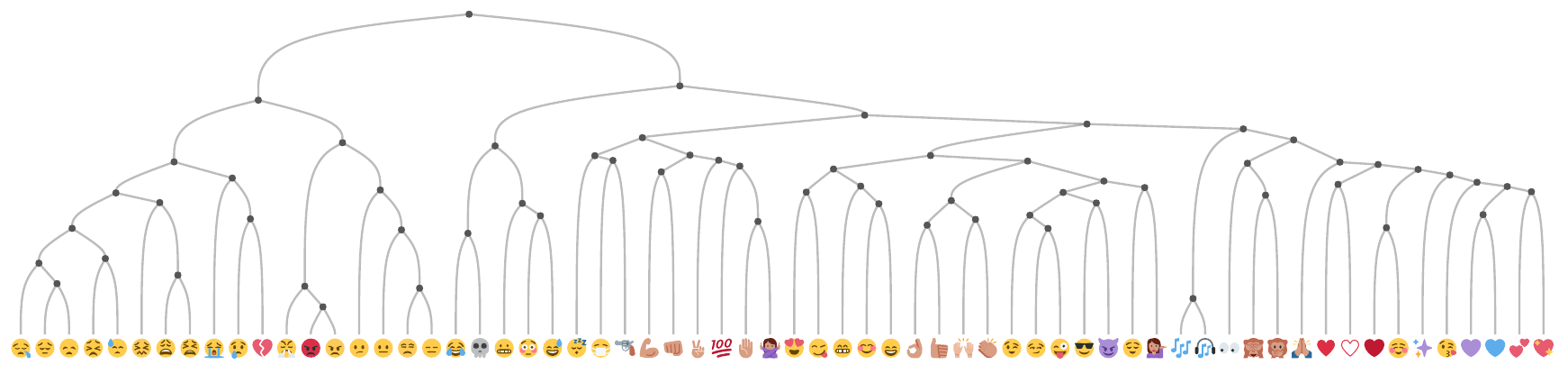} 
  \caption{Hierarchical clustering of the DeepMoji model's predictions across categories on the test set. The dendrogram shows how the model learns to group emojis into overall categories and subcategories based on emotional content. The y-axis is the distance on the correlation matrix of the model's predictions measured using average linkage. More details are available in the supplementary material.}
  \label{fig:tree}
\end{figure*}

One of the major differences between this work compared to previous papers using distant supervision is the diversity of the noisy labels used (see \S\ref{sec:related_work}). For instance, both \citet{deriu2016swisscheese} and \citet{tang2014learning} only used positive and negative emoticons as noisy labels. Other instances of previous work have used slightly more nuanced sets of noisy labels (see \S\ref{sec:related_work}), but to our knowledge our set of noisy labels is the most diverse yet. To analyze the effect of using a diverse emoji set we create a subset of our pretraining data containing tweets with one of 8 emojis that are similar to the positive/negative emoticons used by~\citet{tang2014learning} and \citet{hu2013unsupervised} (the set of emoticons and corresponding emojis are available in the supplemental material). As the dataset based on this reduced set of emojis contains 433M tweets, any difference in performance on benchmark datasets is likely linked to the diversity of labels rather than differences in dataset sizes.

We train our DeepMoji model to predict whether the tweets contain a positive or negative emoji and evaluate this pretrained model across the benchmark datasets. We refer to the model trained on the subset of emojis as \textit{DeepMoji-PosNeg} (as opposed to \textit{DeepMoji}). To test the emotional representations learned by the two pretrained models the `last' transfer learning approach is used for the comparison, thereby only allowing the models to map already learned features to classes in the target dataset. Table~\ref{tab:reduced_emojis} shows that DeepMoji-PosNeg yields lower performance compared to DeepMoji across all \ndsets{}~benchmarks, thereby showing that the diversity of our emoji types encourage the model to learn a richer representation of emotional content in text that is more useful for transfer learning.

\begin{table}[h]
\centering
\small
\caption{Benchmarks using a smaller emoji set (Pos/Neg emojis) or a classic architecture (standard LSTM). Results for DeepMoji from Table~\ref{tab:benchmark_results} are added for convenience. Evaluation metrics are as in Table~\ref{tab:benchmark_results}. Reported values are the averages across five runs.}
\label{tab:reduced_emojis}
\begin{center}
\begin{tabular}{L{1.9cm}C{1.3cm}C{1.3cm}C{1.3cm}}
\toprule
Dataset & Pos/Neg emojis & Standard LSTM & DeepMoji \\
 \midrule
  SE0714  & $.32$ & $.35$ & $.36$ \\ 
  Olympic  & $.55$ & $.57$ & $.61$ \\ 
  PsychExp  & $.40$ & $.49$ & $.56$ \\ 
 \midrule
 SS-Twitter  & $.86$ & $.86$ & $.87$ \\ 
 SS-Youtube  & $.90$ & $.91$ & $.92$ \\ 
 SE1604  & $.56$ & $.57$ & $.58$ \\ 
 \midrule
 SCv1 & $.66$ & $.66$ & $.68$ \\ 
 SCv2-GEN & $.72$ & $.73$ & $.74$ \\ 
\bottomrule
\end{tabular}
\end{center}
\end{table}

Many of the emojis carry similar emotional content, but have subtle differences in usage that our model is able to capture. Through hierarchical clustering on the correlation matrix of the DeepMoji model's predictions on the test set we can see that the model captures many similarities that one would intuitively expect (see Figure~\ref{fig:tree}). For instance, the model groups emojis into overall categories associated with e.g. negativity, positivity or love. Similarly, the model learns to differentiate within these categories, mapping sad emojis in one subcategory of negativity, annoyed in another subcategory and angry in a third one. 

\subsection{Model architecture}
\label{sub_sec:analysis_model_architecture}

Our DeepMoji model architecture as described in \S\ref{sub_sec:model} use an attention mechanism and skip-connections to ease the transfer of the learned representation to new domains and tasks. Here we compare the DeepMoji model architecture to that of a standard 2-layer LSTM, both compared using the `last' transfer learning approach. We use the same regularization and training parameters.

As seen in Table~\ref{tab:reduced_emojis} the DeepMoji model performs better than a standard 2-layer LSTM across all benchmark datasets. The two architectures performed equally on the pretraining task, suggesting that while the DeepMoji model architecture is indeed better for transfer learning, it may not necessarily be better for single supervised classification task with ample available data.

A reasonable conjecture is that the improved transfer learning performance is due to two factors: a) the attention mechanism with skip-connections provide easy access to learned low-level features for any time step, making it easy to use this information if needed for a new task b) the improved gradient-flow from the output layer to the early layers in the network due to skip-connections~\cite{graves2013generating} is important when adjusting parameters in early layers as part of transfer learning to small datasets. Detailed analysis of whether these factors actually explain why our architecture outperform a standard 2-layer LSTM is left for future work.

\subsection{Analyzing the effect of pretraining}
\label{sub_sec:understanding_pretraining}

Performance on the target task benefits strongly from pretraining as shown in Table~\ref{tab:benchmark_results} by comparing DeepMoji (new) to DeepMoji (chain-thaw). In this section we  experimentally decompose the benefit of pretraining into 2 effects: word coverage and phrase coverage. These two effects help regularize the model by preventing overfitting (see the supplementary details for an visualization of the effect of this regularization).

There are numerous ways to express a specific sentiment, emotion or sarcastic comment. Consequently, the test set may contain specific language use not present in the training set. The pretraining helps the target task models attend to low-support evidence by having previously observed similar usage in the pretraining dataset. We first examine this effect by measuring the improvement in word coverage on the test set when using the pretraining with word coverage being defined as the \% of words in the test dataset seen in the training/pretraining dataset (see Table~\ref{tab:word_coverage}). An important reason why the `chain-thaw' approach outperforms other transfer learning approaches is can be used to tune the embedding layer with limited risk of overfitting. Table~\ref{tab:word_coverage} shows the increased word coverage from adding new words to the vocabulary as part of that tuning.

Note that word coverage can be a misleading metric in this context as for many of these small datasets a word will often occur only once in the training set. In contrast, all of the words in the pretraining vocabulary are present in thousands (if not millions) of observations in the emoji pretraining dataset thus making it possible for the model to learn a good representation of the emotional and semantic meaning. The added benefit of pretraining for learning word representations therefore likely extends beyond the differences seen in Table~\ref{tab:word_coverage}.

\begin{table}[h]
\centering
\small
\caption{Word coverage on benchmark test sets using only the vocabulary generated by finding words in the training data (`own'), the pretraining vocabulary (`last') or a combination of both vocabularies (`full / chain-thaw').}
\label{tab:word_coverage}
\begin{center}
\begin{tabular}{L{1.6cm}C{1.5cm}C{1.5cm}C{1.5cm}}
\toprule
Dataset & Own & Last & Full / Chain-thaw \\
 \midrule
  SE0714  & $41.9$\% & $93.6$\% & $94.0$\% \\
  Olympic  & $73.9$\% & $90.3$\% & $96.0$\% \\
  PsychExp  & $85.4$\% & $98.5$\% & $98.8$\% \\
 \midrule
 SS-Twitter  & $80.1$\% & $97.1$\% & $97.2$\% \\
 SS-Youtube  & $79.6$\% & $97.2$\% & $97.3$\% \\
 SE1604  & $86.1$\% & $96.6$\% & $97.0$\% \\
 \midrule
 SCv1 & $88.7$\% & $97.3$\% & $98.0$\% \\
 SCv2-GEN & $86.5$\% & $97.2$\% & $98.0$\% \\
\bottomrule
\end{tabular}
\end{center}
\end{table}

To examine the importance of capturing phrases and the context of each word, we evaluate the accuracy on the SS-Youtube dataset using a fastText classifier pretrained on the same emoji dataset as our DeepMoji model. This fastText classifier is almost identical to only using the embedding layer from the DeepMoji model. We evaluate the representations learned by fine-tuning the models as feature extractors (i.e. using the `last' transfer learning approach). The fastText model achieves an accuracy of $63$\% as compared to $93$\% for our DeepMoji model, thereby emphasizing the importance of phrase coverage. One concept that the LSTM layers likely learn is negation, which is known to be important for sentiment analysis~\cite{wiegand2010survey}.

\subsection{Comparing with human-level agreement}
\label{sub_sec:comparing_humans}

To understand how well our DeepMoji classifier performs compared to humans, we created a new dataset of random tweets annotated for sentiment. Each tweet was annotated by a minimum of 10 English-speaking Amazon Mechanical Turkers (MTurk's) living in USA. Tweets were rated on a scale from 1 to 9 with a `Do not know' option, and guidelines regarding how to rate the tweets were provided to the human raters. The tweets were selected to contain only English text, no mentions and no URL's to make it possible to rate them without any additional contextual information. Tweets where more than half of the evaluators chose `Do not know' were removed (98 tweets).

For each tweet, we select a MTurk rating random to be the `human evaluation', and average over the remaining nine MTurk ratings are averaged to form the ground truth. The `sentiment label' for a given tweet is thus defined as the overall consensus among raters (excluding the randomly-selected `human evaluation' rating). To ensure that the label categories are clearly separated, we removed neutral tweets in the interval $[4.5, 5.5]$ (roughly $29\%$ of the tweets). The remaining dataset consists of $7\,347$ tweets. Of these tweets, $5000$ are used for training/validation and the remaining are used as the test set. Our DeepMoji model is trained using the chain-thaw transfer learning approach.

Table~\ref{tab:mturk_results} shows that the agreement of the random MTurk rater is $76.1\%$, meaning that the randomly selected rater will agree with the average of the nine other MTurk-ratings of the tweet's polarity $76.1\%$ of the time. Our DeepMoji model achieves $82.4\%$ agreement, which means it is better at capturing the average human sentiment-rating than a single MTurk rater. 

\begin{table}[th]
\centering
\small
\caption{Comparison of agreement between classifiers and the aggregate opinion of Amazon Mechanical Turkers on sentiment prediction of tweets.}
\label{tab:mturk_results}
\begin{center}
\begin{tabular}{@{\hspace{3pt}}l@{\hspace{3pt}}c}
\toprule
\hspace{4cm} & Agreement \\
 \midrule
 Random  & $50.1\%$ \\
 fastText & $71.0\%$ \\
 MTurk & $76.1\%$ \\ 
 DeepMoji & $\mathbf{82.4}\%$ \\
\bottomrule
\end{tabular}
\end{center}
\end{table}

\section{Conclusion}
\label{sec:discussion}

We have shown how the millions of texts on social media with emojis can be used for pretraining models, thereby allowing them to learn representations of emotional content in texts. Through comparison with an identical model pretrained on a subset of emojis, we find that the diversity of our emoji set is important for the performance of our method. We release our pretrained DeepMoji model with the hope that other researchers will find good use of them for various emotion-related NLP tasks\footnote{Available with preprocessing code, examples of usage, benchmark datasets etc. at github.com/bfelbo/deepmoji}.

\section*{Acknowledgments}

The authors would like to thank Janys Analytics for generously allowing us to use their dataset of human-rated tweets and the associated code to analyze it. Furthermore, we would like to thank Max Lever, who helped design the online demo, and Han Thi Nguyen, who helped code the software that is provided alongside the pretrained model.

\bibliography{emnlp2017}
\bibliographystyle{emnlp_natbib}

\appendix

\section{Supplemental Material}
\label{sec:supplemental}

\subsection{Preprocessing Emotion Datasets}
\label{sub_sec:supp_emotion_dsets}

In the Olympic Games dataset by \citeauthor{sintsova2013fine} each tweet can be assigned multiple emotions out of 20 possible emotions, making evaluation difficult. To counter this difficulty, we have chosen to convert the labels to 4 classes of low/high valence and low/high arousal based on the Geneva Emotion Wheel that the study used. A tweet is deemed as having emotions within the valence/arousal class if the average evaluation by raters for that class is $2.0$ or higher, where `Low' = $1$, `Medium' = $2$ and `High' = $3$.

We also evaluate on the ISEAR databank~\cite{wallbott1986universal}, which was created over many years by a large group of psychologists that interviewed respondents in 37 countries. Each observation in the dataset is a self-reported experience mapped to 1 of 7 possible emotions, making for an interesting benchmark dataset.

\subsection{Pretraining as Regularization}

\begin{figure}[hpt]
  \centering
  \includegraphics[trim=0.0cm 0.0cm 0.0cm 0.0cm, clip, width=1\columnwidth]{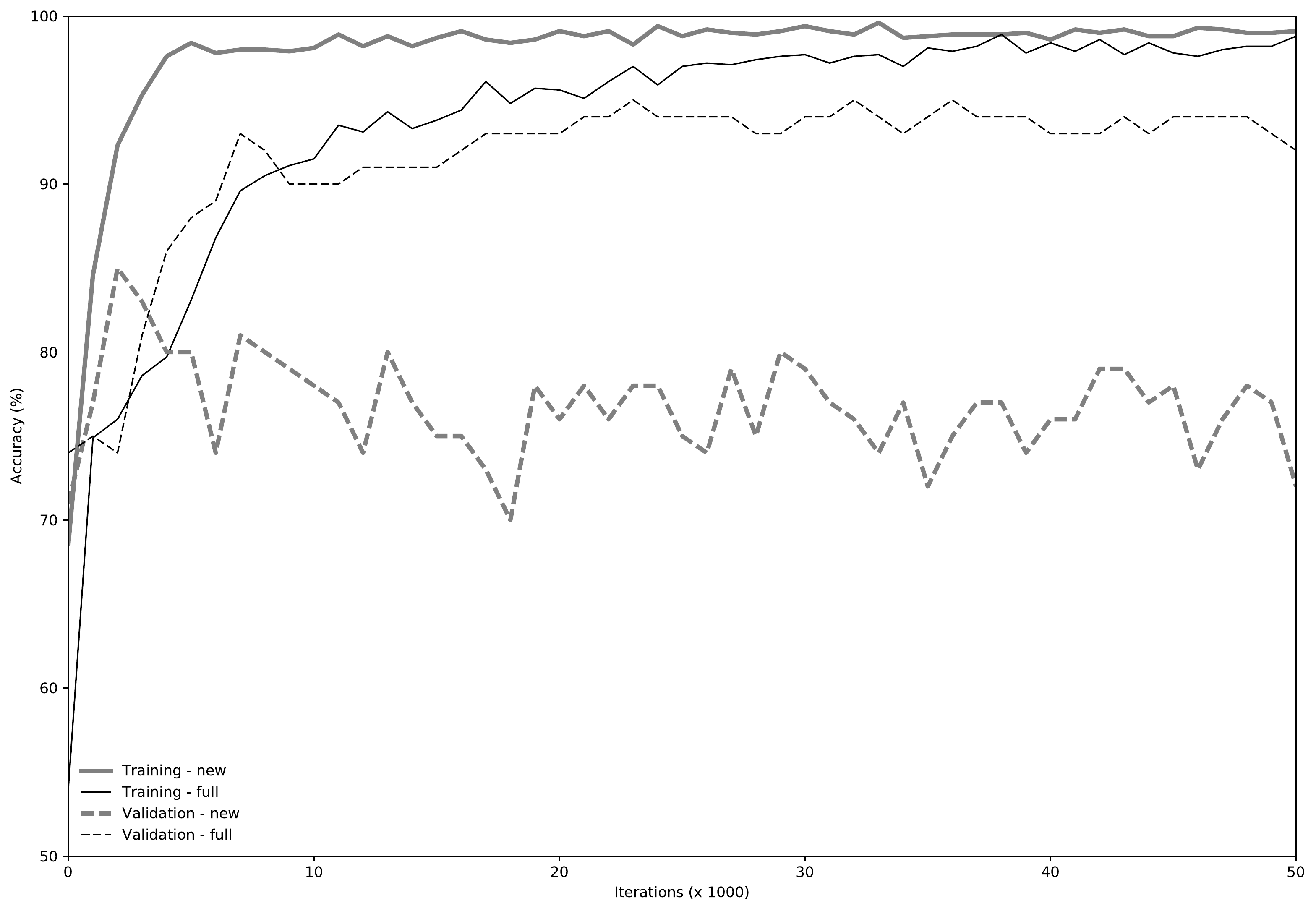}
  \caption{Training statistics on the SS-Youtube dataset with a pretrained model vs. a untrained model. The architecture and all hyperparameters are identical for the two models. All layers are unfrozen.}
  \label{fig:pretraining_regularization}
\end{figure}

Figure~\ref{fig:pretraining_regularization} shows an example of how the pretraining helps to regularize the target task model, which otherwise quickly overfits. The chain-thaw transfer learning approach further increases this regularization by fine-tuning the model layer wise, thereby adding additional regularization.

\subsection{Emoticon to Emoji mapping}

\begin{figure}[hpt]
  \centering
  \includegraphics[trim=0.0cm 0.0cm 0.0cm 0.0cm, clip, width=0.25\columnwidth]{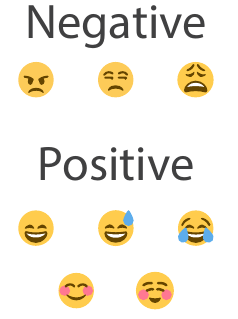}
  \caption{Emojis used for the experiment on the importance of a diverse noisy label set.}
  \label{fig:pos_neg_overview}
\end{figure}

To analyze the effect of using a diverse emoji set we create a subset of our pretraining data containing tweets with one of 8 emojis that are similar to the positive/negative emoticons used by~\citet{tang2014learning} and \citet{hu2013unsupervised}. The positive emoticons are :) : ) :-) :D =) and the negative emoticons are :( : ( :-(. We find the 8 similar emojis in our dataset seen in Figure~\ref{fig:pos_neg_overview} as use these for creating the reduced subset.

\subsection{Emoji Clustering}

\begin{figure*}[thp]
  \centering
  \includegraphics[trim=0.0cm 0.0cm 0.0cm 0.0cm, clip, width=1\textwidth]{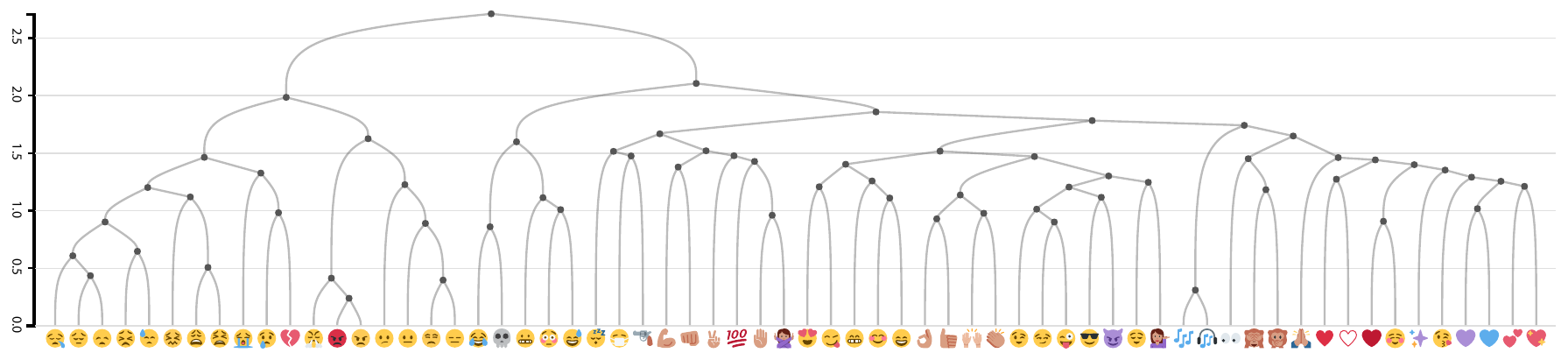}
  \caption{Hierarchical clustering of the DeepMoji model's predictions across categories on the test set. The dendrogram shows how the model learns to group emojis into overall categories and subcategories based on emotional content. The y-axis is the distance on the correlation matrix of the model's predictions measured using average linkage.}
  \label{fig:tree_scale}
\end{figure*}

\begin{figure*}[htp]
  \centering
  \includegraphics[trim=0.0cm 0.0cm 0.0cm 0.0cm, clip, width=1\textwidth]{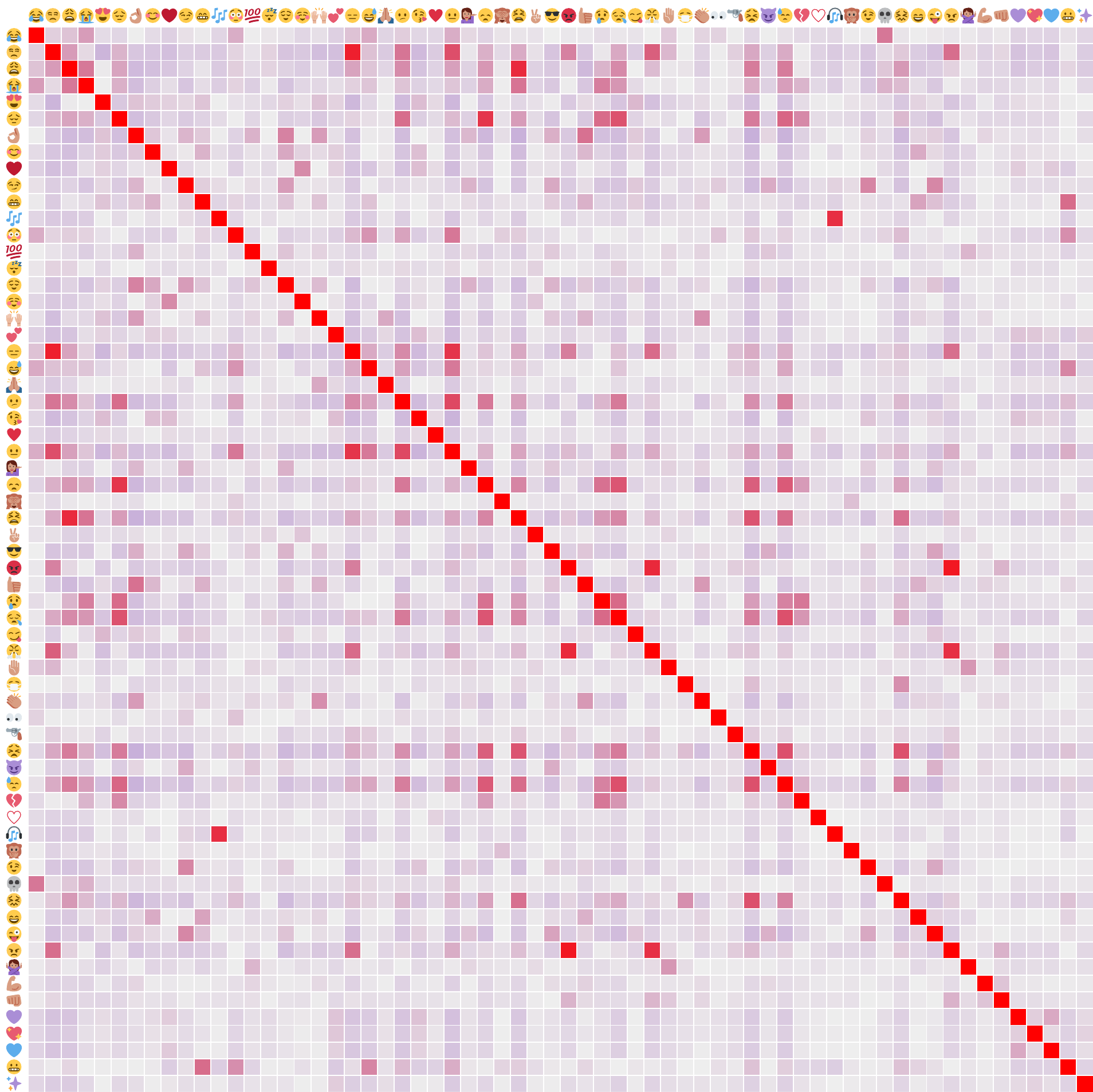}
  \caption{Correlation matrix of the model's predictions on the pretraining test set.}
  \label{fig:corr_matrix}
\end{figure*}

We compute the predictions of the DeepMoji model on the pretraining test set containing $640$K tweets and compute the correlation matrix of the predicted probabilities seen in Figure~\ref{fig:corr_matrix}. Then we use hierarchical clustering with average linkage on the correlation matrix to generate the dendrogram seen in Figure~\ref{fig:tree_scale}. We visualized dendrograms for various versions of our model and the overall structure is very stable with only a few emojis changing places in the hierarchy.

\end{document}